\pdfoutput=1

\documentclass[11pt]{article}
\usepackage{soul}
\usepackage[]{acl}

\usepackage{times}
\usepackage{latexsym}
\usepackage{setspace}

\usepackage[T1]{fontenc}

\usepackage[utf8]{inputenc}

\usepackage{microtype}

\usepackage{graphicx}
\usepackage{booktabs}
\usepackage{multirow}
\usepackage[normalem]{ulem}
\usepackage{adjustbox}
\usepackage{tcolorbox}

\useunder{\uline}{\ul}{}
\usepackage{pifont}
\newcommand{\NO}{\ding{51}}
\newcommand{\YES}{\ding{55}}
\newcommand{\cmmnt}[1]{}

\usepackage{enumitem}
\usepackage{linguex}
\usepackage{lipsum}

\usepackage{booktabs}
\usepackage{longtable}
\usepackage{array}
\usepackage{multirow}
\usepackage{wrapfig}
\usepackage{float}
\usepackage{colortbl}
\usepackage{pdflscape}
\usepackage{tabu}
\usepackage{threeparttable}
\usepackage{threeparttablex}
\usepackage[normalem]{ulem}
\usepackage{makecell}
\usepackage{xcolor}
\usepackage[toc,page]{appendix}

\title{Assessing the influence of attractor-verb distance on grammatical agreement in humans and language models}

\author{Christos-Nikolaos Zacharopoulos \\
  Cognitive Neuroimaging Unit,\\NeuroSpin center, France\\
  Sensome SAS,\\ Massy,\\ France\\
  \texttt{\footnotesize {christonik@gmail.com}}
  \And
  Théo Desbordes \\
  Meta AI Research;\\
  Cognitive Neuroimaging Unit,\\NeuroSpin center, France;\\
  \texttt{\footnotesize desbordes.theo@gmail.com} \\
  \And
  Mathias Sablé-Meyer\\
  Cognitive Neuroimaging Unit,\\ NeuroSpin center, France;\\
  Collège de France,\\Université PSL\\
  \texttt{\footnotesize mathias.sable-meyer@ens-cachan.fr}
  }

\begin{document}

\maketitle

\begin{abstract}
  Subject-verb agreement in the presence of an attractor noun located between
  the main noun and the verb elicits complex behavior: judgments of
  grammaticality are modulated by the grammatical features of the attractor.
  For example, in the sentence \textit{“The girl near the boys likes
  climbing''},  the attractor (\emph{boys}) disagrees in grammatical number
  with the verb (\emph{likes}), creating a locally implausible transition
  probability. Here, we parametrically modulate the distance between the
  attractor and the verb while keeping the length of the sentence equal. We
  evaluate the performance of both humans and two artificial neural network
  models: both make more mistakes when the attractor is closer to the verb, but
  neural networks get close to the chance level while humans are mostly able to
  overcome the attractor interference. Additionally, we report a linear effect
  of attractor distance on reaction times. We hypothesize that a possible
  reason for the proximity effect is the calculation of transition
  probabilities between adjacent words. Nevertheless, classical models of
  attraction such as the cue-based model might suffice to explain this
  phenomenon, thus paving the way for new research. Data and analyses available
  at  {\footnotesize \url{https://osf.io/d4g6k}}
\end{abstract}

\section{Introduction}

On the surface, language appears to be produced and understood linearly, as
humans read or hear words one after the other. Yet, formal linguistic theories
postulate the existence of an underlying structure that governs language
processing \cite{chomsky1957, rizzi2004cartography, dehaene2015neural,
vigliocco1998separating, hauser2002faculty}. This hypothesis is supported by
both behavioral \cite{fossum-levy-2012-sequential,
shi2020toddlers,coopmans2021structure} and neural \cite{brennan_abstract_2016,
nelson2017neurophysiological, pallier2011cortical} data. According to a
competing view, unstructured probabilistic models capture behavior without
explicitly relying on structures \cite{frank2011insensitivity}. The discrepancy
between the two views has led to a decade-long debate on linear versus
structural effects in language \cite{haskell2005constituent, Ding2015,
willer2017linearity, arana2021mvpa}.

\begin{figure}
  \centering
  \includegraphics{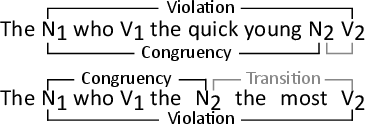}
  \caption{\textbf{Experimental design}
    Our design aims to separate two possible mechanisms involved in language
    processing, by altering the distance of the embedded noun (attractor) from
    the verb. We compare the impact of noun-verb agreement (\emph{Violation})
    and noun-noun dependency (\emph{Congruency}). The \emph{Transition} effect
    symbolizes the consequence of the transition probability between the second
    noun and the verb. As a baseline, we also include a condition with a
    similar number of words, but without a noun in between. The summary of the
    experimental conditions is shown in Table \ref{tab:table1}}
    \label{fig:pedagogical}
\end{figure}

Analyzing subject-verb agreement errors in the presence of distracting elements
is a standard way of separating linear and structural accounts of language
\cite{molinaro2011grammatical}. Attraction effects in number agreement have
been studied extensively in humans \cite{bock1991broken, franck2002subject,
hammerly2019grammaticality} and Neural Language Models (NLMs)
\cite{linzen2016assessing,  jumelet2019analysing, lakretz2021rnns}.

\renewcommand{\arraystretch}{0.5}
\begin{table*}[t!]
  \centering
  \begin{footnotesize}
    \begin{adjustbox}{width=\linewidth}
      \begin{tabular}{@{}clccc@{}}
        \toprule
        Condition                           & \multicolumn{1}{c}{Sentence}                             & Violation & Congruency \\ \midrule
        \multirow{4}{*}{Proximal Attractor} & The writer who knows the happy young journalist climbs.  & \YES      & \NO        \\ \
                                            & The writer who knows the happy young journalists climbs. & \YES      & \YES       \\
                                            & The writer who knows the happy young journalist climb.   & \NO       & \NO        \\ \
                                            & The writer who knows the happy young journalists climb.  & \NO       & \YES       \\ \midrule
        \multirow{4}{*}{Distal Attractor}   & The writer who knows the journalist the most climbs.     & \YES      & \NO        \\
                                            & The writer who knows the journalists the most climbs.    & \YES      & \YES       \\
                                            & The writer who knows the journalist the most climb.      & \NO       & \NO        \\
                                            & The writer who knows the journalists the most climb.     & \NO       & \YES       \\ \midrule
        \multirow{2}{*}{Baseline}           & The writer who walks fast but rather clumsily climbs.    & \YES      & -          \\
                                            & The writer who walks fast but rather clumsily climb.     & \NO       & -          \\ \midrule
        \multirow{2}{*}{Filler (Number)}    & The writer who knows the journalist climbs.              & \YES      & -          \\
                                            & The writer who know the journalist climbs.               & \NO       & -          \\ \midrule
        \multirow{2}{*}{Filler (POS)}       & The writer whom the journalist knows climbs.             & \YES      & -          \\
                                            & The writer whom knows the journalist climbs.             & \NO       & -          \\ \bottomrule
      \end{tabular}
    \end{adjustbox}
    \caption{\textbf{Conditions.} The design contrasts the attribution of two
    distinct effects to the overall error rate in humans and NLMs. To prevent
    the subjects from developing strategies for the effective resolution of the
    violation-detection task, we included two different kinds of filler trials
    that contained violations at the beginning of the sentence.}
    \label{tab:table1}
  \end{footnotesize}
\end{table*}

In the present work, we introduce a parametric manipulation of the distance
between the distracting nouns and the verb (Figure~\ref{fig:pedagogical}). We
include a baseline condition where the subject-verb distance is matched but no
word that carries number marking is introduced in-between. We thereby provide a
minimal triplet of experimental conditions (Table~\ref{tab:table1}) that can
disentangle structural from linear mechanisms by contrasting operations
directly ascribed to each mechanism. We posit that there are structural
operations at play to explain the fact that participants are always able to
detect violations, and at the same time there are linear operations at play
which lead to the attraction effect of the intervening noun. We additionally
hypothesize that the linear effect might be modulated by transition
probabilities, i.e., the prior probability that a given word follows another
\cite{dehaene2015neural,friston2021}. Finally, despite reaching on par with or
even supra-human performance on many tasks \cite{brown2020gpt3,
minaee2021deep}, NLMs are known to be sensitive to superficial statistical
properties of their training data \cite{mccoy2021much}. Thus, we compare two
NLMs with the behavior of human participants.

\section{Experimental Evidence}

\subsection{Method}

We tested the grammatical judgments of the participants in a forced-choice,
online violation detection task where the words were presented one at a time on
the screen (RSVP), and the participants had to press a button to indicate
whether a given sentence was grammatically correct or not.

\paragraph{Participants}

Fifty-four native speakers of English took part in our experiment, which was
advertised on social media and mailing lists. The procedure and the consent
were approved by the local ethical committee (Université Paris-Saclay,
CER-Paris-Saclay-2019-063). We used filler trials (Table 1) to avoid potential
confounding strategies from the participants, such as actively ignoring the
middle of the sentences \cite{pearlmutter1999agreement}. Any participant whose
answer to fillers was not significantly different from chance (binomial test,
null hypothesis $p_0=.5$) was rejected. We also rejected participants whose
success rate on the main task was below $70\%$ Overall, we rejected 20
participants, and reported analyses from 34 participants (all analyses reported
are consistent with corresponding analyses performed with the full dataset,
reported in appendix \ref{appendix:appendixB}. For example, one can compare
Table~\ref{tab:table2} and Table~\ref{tab:table2fulldataset}, and Figure
\ref{fig:aggregatedResults} and Figure~\ref{fig:fig2allparticipants}).

We also tested two transformer models \cite{wolf2020huggingfaces}: a
replication of the GPT-3 language model \cite{brown2020gpt3} made available by
EleutherAI\footnote{\tiny\url{https://huggingface.co/EleutherAI/gpt-neo-1.3B}}
\cite{gpt-neo} and a Text-To-Text Transformer
(T5)\footnote{\tiny\url{https://huggingface.co/vennify/t5-base-grammar-correction}}
\cite{raffel2019t5} fine-tuned on a grammatical error correction benchmark
\cite{napoles2017jfleg}. To evaluate the GPT-3 model, we input it with the
sentence up to (excluding) the target verb and compare the probabilities
associated with the grammatical and ungrammatical tokens (e.g., “climb'' vs
“climbs'' for sentences in Table~1). Thus, for this model, we get a comparative
performance per condition but cannot evaluate performances between grammatical
and ungrammatical sentences. On the other hand, we compare humans directly with
the grammaticality judgment of T5.

\paragraph{Experimental Procedure}

Participants were given a description of a violation-detection task (see
Appendix \ref{appendix:appendixA} for the exact wording) including which button
to press and three examples of grammatical and ungrammatical sentences. We
presented sentences to participants, with a fixation cross between words, in
white on black background, using a presentation time of 200ms and an SOA of
366ms. Participants could answer at any point by pressing the keyboard to
indicate their judgment of grammaticality (left and right arrow keys randomized
across subjects). Participants received auditory and visual feedback with each
trial: green/red fixation and upward/downward tune (correct/incorect). At the
end of the experiment, participants answered questions about the experiment
(difficulty, weirdness, strategies), we provided them with an overall score and
invited them to share the experiment on social media.

\paragraph{Stimuli}

We generated sentences from a fixed lexicon balanced for low-level features,
yielding many sentences. Then we filtered them based on the perplexity of
GPT-3, keeping only sentences that had an overall perplexity between the median
and median $+ 2 std$ --- in order to keep sentences of low and consistent
perplexity. We sampled $5$ sentences for each condition (baseline, distal
congruent, distal incongruent, proximal congruent, proximal incongruent),
subject number (singular, plural), and grammaticality, as well as 32 filler
sentences. No two sentences shared 5 words or more. The same sentences were
presented to participants in randomized order.

\subsection{Results}

\begin{figure}
  \centering
  \includegraphics{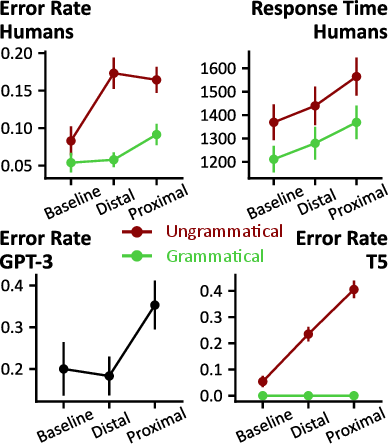}
  \caption{Performances of humans and NLMs: colors indicate grammaticality,
  error bars indicate (all figures) SEM over participants (humans) or sentences
  (NLMs).}
  \label{fig:aggregatedResults}
\end{figure}

We call \emph{incongruent} the trials in which the numbers of $N_1$ and $N_2$
disagree (Figure \ref{fig:pedagogical}) --- which is independent from
\emph{grammaticality}. Figure \ref{fig:aggregatedResults} shows the main effect
of the presence of an attractor, and how its distance to the target (distal or
proximal) modulates the performances of humans and NLMs.

In all cases, the baseline elicits fewer errors (ER) and faster reaction times
(RT) compared to the attractor conditions.  Errors occur more often in
ungrammatical sentences than in grammatical ones, irrespective of the
condition: this phenomenon is called \textit{grammatical illusions}
\cite{phillips2011grammatical} and indicates that participants accept
ungrammatical sentences more often than they reject grammatical ones. We also
replicate \textit{grammatical asymmetry} \cite{wagers2009agreement}:
incongruent trials lead to higher ER in the ungrammatical sentences.

To investigate the attractor effects, we analyzed the main factors of our
design. The main effect of \textit{Violation} refers to the dependency that
controls the grammatical configuration of the sentence
(Figure~\ref{fig:pedagogical}) and we use this effect as a proxy into
structural processing \cite{rizzi2004cartography}. The \textit{Congruency}
effect refers to a dependency realized between two non-structurally related
words and is used as a proxy for linear processing.

\begin{figure*}
  \centering
  \includegraphics[width=0.8\textwidth]{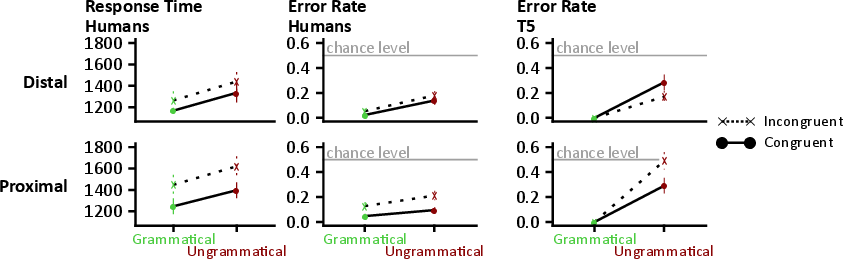}
  \caption{Effect of grammaticality, congruency, and distance of the attractor
  on our dependent variables.}
  \label{fig:detailedResults}
\end{figure*}

Figure~\ref{fig:detailedResults} shows the main factors and their interaction.
We report the corresponding ANOVA for the human participants in
Table~\ref{tab:table2}. We replicate the markedness phenomenon
\cite{bock1991broken,wagers2009agreement}: attraction effects surface only with
plural attractors (see Table~\ref{tab:table2singular} and
Figure~\ref{fig:detailedResultsAppendix} for analyses with singular
attractors). Results shown in Figure~\ref{fig:detailedResults} and
Table~\ref{tab:table2singular} correspond to sentences with plural attractors.

\renewcommand{\arraystretch}{0.9}
\begin{table}[t!]
  \footnotesize
  \centering
  \begin{tabular}{lcrr}
    \toprule
    Effect                                                & $F_{1,33}$                             & $p$-value                              & $\eta_G^2$                            \\
    \midrule
    \addlinespace[0.3em]
    \multicolumn{4}{l}{\textbf{Response Time; Distal attractor}}                                                                                                                    \\
    \cellcolor[HTML]{fff3e0}{\em{\hspace{1em}congruency}} & \cellcolor[HTML]{fff3e0}{\em{$7.04$}}  & \cellcolor[HTML]{fff3e0}{\em{$.012$}}  & \cellcolor[HTML]{fff3e0}{\em{$.011$}} \\
    \cellcolor[HTML]{fff3e0}{\em{\hspace{1em}violation}}  & \cellcolor[HTML]{fff3e0}{\em{$11.64$}} & \cellcolor[HTML]{fff3e0}{\em{$.002$}}  & \cellcolor[HTML]{fff3e0}{\em{$.031$}} \\
    \hspace{1em}interaction                               & $0.01$                                 & $.915$                                 & $<.001$                               \\
    \addlinespace[0.3em]
    \multicolumn{4}{l}{\textbf{Response Time; Proximal attractor}}                                                                                                                  \\
    \cellcolor[HTML]{fff3e0}{\em{\hspace{1em}congruency}} & \cellcolor[HTML]{fff3e0}{\em{$26.90$}} & \cellcolor[HTML]{fff3e0}{\em{$<.001$}} & \cellcolor[HTML]{fff3e0}{\em{$.046$}} \\
    \cellcolor[HTML]{fff3e0}{\em{\hspace{1em}violation}}  & \cellcolor[HTML]{fff3e0}{\em{$11.81$}} & \cellcolor[HTML]{fff3e0}{\em{$.002$}}  & \cellcolor[HTML]{fff3e0}{\em{$.027$}} \\
    \hspace{1em}interaction                               & $0.10$                                 & $.752$                                 & $<.001$                               \\
    \addlinespace[0.3em]
    \multicolumn{4}{l}{\textbf{Error Rate; Distal attractor}}                                                                                                                       \\
    \hspace{1em}congruency                                & $2.29$                                 & $.140$                                 & $.013$                                \\
    \cellcolor[HTML]{fff3e0}{\em{\hspace{1em}violation}}  & \cellcolor[HTML]{fff3e0}{\em{$18.01$}} & \cellcolor[HTML]{fff3e0}{\em{$<.001$}} & \cellcolor[HTML]{fff3e0}{\em{$.135$}} \\
    \hspace{1em}interaction                               & $0.04$                                 & $.846$                                 & $<.001$                               \\
    \addlinespace[0.3em]
    \multicolumn{4}{l}{\textbf{Error Rate; Proximal attractor}}                                                                                                                     \\
    \cellcolor[HTML]{fff3e0}{\em{\hspace{1em}congruency}} & \cellcolor[HTML]{fff3e0}{\em{$15.79$}} & \cellcolor[HTML]{fff3e0}{\em{$<.001$}} & \cellcolor[HTML]{fff3e0}{\em{$.085$}} \\
    \cellcolor[HTML]{fff3e0}{\em{\hspace{1em}violation}}  & \cellcolor[HTML]{fff3e0}{\em{$5.33$}}  & \cellcolor[HTML]{fff3e0}{\em{$.027$}}  & \cellcolor[HTML]{fff3e0}{\em{$.040$}} \\
    \hspace{1em}interaction                               & $0.57$                                 & $.454$                                 & $.003$                                \\
    \bottomrule
  \end{tabular}
  \caption{One-way between-subjects ANOVAs were conducted to compare the effect
  of congruency, violation, and their interaction on RT and ER for both
  proximal and distal attractors. Highlighted rows indicate significance at the
  $p<.05$ level. The last column provides $\eta_G^2$ values, an estimator of
  the variance explained by the ANOVA, similar to $r^2$ for linear models.}
  \label{tab:table2}
\end{table}

In the human data, the main effect of \textit{Violation} is significant across
all conditions and dependent variables. Nevertheless, for the ER, the
$\eta_G^2$ value is larger by an order of magnitude in the distal attractor
condition: this illustrates that participants make more mistakes in judging the
grammaticality of sentences in which the attractor is far from the verb,
especially when the sentence is ungrammatical. The distance of the attractor
does not affect the magnitude of the effect in the case of RTs.

To elucidate this facilitatory effect, we focus on \textit{Congruency}: its
effect is only significant in the proximal condition, which illustrates that
participants make more errors in incongruent sentences compared to the
congruent ones (figure \ref{fig:detailedResults}).

NLMs displayed super-human performances in evaluating grammatical sentences. T5
is sensitive to the mere presence of an attractor, whose distance strongly
modulates performance. GPT-3 is not sensitive to the mere presence of an
attractor, but elicits a significant distance effect. Both models, but not
humans, have near-chance performances in the case of a proximal attractor.
These results might have a two-fold interpretation. The presence of
\textit{grammatical illusions} in both humans and models might be informative
on the role of training in linguistic performance: in real sentences,
grammatical sentences vastly outnumber ungrammatical ones. We might therefore
be describing a similar training bias between humans and NLMs. There is a
common response profile between the NLMs, which demonstrates the sensitivity of
transformer models to statistical regularities. This allows us to zoom in on
T5, which is more directly comparable to humans, and investigate how the
\textit{Congruency} factor modulates the ER.

The reduction in the ER in the human data was traced to a facilitatory effect
of congruency: congruent trials led to fewer errors in the proximal compared to
the distal attractor. On the contrary, in T5, the incongruent trials yielded
chance-level performances, but there was no distance effect in the congruent
trials. This indicates that congruent sentences help humans in performing the
task, but not NLMs. Additionally, incongruent sentences have a detrimental
effect in the models, but no such effect is observed in humans. These
observations point to a common sensitivity to the attractor-verb distance in
both systems, but a fundamental difference as to the outcome of this
sensitivity.

\section{Discussion}

In this study, we revisited the classical attraction effect in subject-verb
number agreement in humans and neural networks and sought to assess the
influence of the attractor-verb distance.

Our results draw a picture of a shared distance sensitivity between humans and
NLMs, but also a fundamental difference in the weight of this sensitivity. We
observed that the artificial system operates on the basis of word-level
statistics, and is thus driven to chance-level performance in the presence of a
deviant bigram (Figure~\ref{fig:detailedResults}-bottom right-Incongruent
trials). On the contrary, the incongruency of the sentence leads to comparable
error rates irrespective of the attractor distance in humans
(Figure~\ref{fig:detailedResults}-central column). The significant effect of
congruency observed in the human data is due to a facilitatory effect of
congruency in judging grammaticality, and not an inhibitory effect of
incongruency, unlike with the neural networks. This effect is mostly evident in
the agrammatical sentences and can be decomposed as follows.

Consider the following two sentences:

\begin{itemize}
  \item [$(a)$] [\textit{Congruent}]\newline The writers who like the happy young \newline \textbf{editors cries}.
  \item [$(b)$] [\textit{Incongruent}]\newline The writers who like the happy young \newline \textbf{editor cries}.
\end{itemize}

Participants and $T5$ made fewer errors in $(a)$ compared to $(b)$.
Participants also made fewer errors in $(a)$ when the attractor was further
away. We thus observe a facilitatory effect. Notice that here we have the
realization of the locally implausible bigram \textbf{“editors cries”}. One
possible interpretation might be that the participants were lured into judging
this sentence as invalid, based on the presence of this bigram. A facilitatory
effect is hence observed because the sentence was indeed agrammatical.

The error rate for the incongruent cases like $(b)$ remained unchanged for the
humans, with respect to the distal attractor (see
Figure~\ref{fig:detailedResults}-central column). Importantly, though, in this
case, where a plausible bigram is found (\textbf{“editor cries”}), the NLM
reached chance-level performance. Given the local agreement of the attractor
with the verb, the NLM was lured into judging the sentence as grammatical, when
the sentence was wrong, whereas humans were able to mostly overcome the
attraction effect.

Thus, our results suggest that operations at the n-gram level might play a key
role in explaining the observed phenomena, in both humans and NLMs.

Nevertheless, alternative explanations can be described. In many behavioral
studies, significant effects of congruency have been reported in
non-intervening attractor structures such as Object Relative Clauses (e.g.: The
player that the [teacher encourage/s] climbs) \cite{wagers2009agreement}. The
n-gram mechanism cannot explain attraction phenomena in this setup, something
that the dominant model of attraction (cue-retrieval model,
\cite{wagers2009agreement}) can do. In this model, a memory mechanism is
enabled upon a cue, that retrieves the number of the subject from the memory
system and the errors can be attributed to retrieval interference. Under this
interpretation, the memory representation attractors fade with distance and
therefore the distal attractor does not compete for retrieval. In contrast,
when the attractor is in the vicinity of the verb, similarity-based retrieval
interference can occur, and thus, attraction effects can only be realized in
that condition \cite{mcelree2003memory}.

However, it is important to note that none of the dominant models of
grammatical agreement (cue-based retrieval model \& feature percolation) are
complete, as both are conclusive in ungrammatical conditions, but not always
for the grammatical ones. Thus, a need of model revision seems necessary. In
this study, we tried to point to the calculation of transition probabilities as
a candidate factor for model development.

\subsection{Limitations}

There are a number of limitations that narrows the scope of our results. First,
it is difficult to draw general conclusions from a single experiment on
grammatical number agreement. Second, although all conditions are balanced, the
stimuli we used are not devoid of semantic content which might induce some
biases. These results will need to be confirmed in future experiments using
different tasks and stimuli, for example using morphosyntactically marked
stimuli but devoid of semantics, so-called “jabberwocky” sentences
\citep{hahne_whats_2001, desbordes_dimensionality_2023}. This subject
represents a significant area for our ongoing research and exploration.

Third, we compared human participants to language models under the assumption
that NLMs express sensitivity to probabilistic relationships at the word level,
and thus a comparison under the same conditions might shed light on the
processing of language in the human brain. We are fully aware that this
comparison is indirect, and that an LSTM architecture might have been more
appropriate for such a comparison. Nevertheless, the literature has pointed to
differences between grammatical and ungrammatical conditions, effects that we
replicated in this study, and we therefore sought for a model that would allow
us for a direct comparison for each condition.

\subsection{Conclusion}

Our results show that, in humans and NLMs, language processing is affected by
the attractor-verb distance. We additionally hypothesize that this is due to
the calculation of transition probabilities at the word level, which can either
run contrary or reinforce the overall structure-based processing.

Humans are less affected by this local interference, suggesting that language
models process language in ways that are still fundamentally different from
humans, even though they superficially coincide, e.g. in grammatical cases in
our data.

\newpage{}

\bibliography{custom}
\bibliographystyle{acl_natbib}

\newpage
\begin{appendices}

  \section{Questionnaire, Instructions, and Stimuli}
  \label{appendix:appendixA}

  \section*{Instructions}

  The exact instructions given to the participants are provided below. They consisted of three separate pages, participants could go back and forth between pages freely.

  Remember that the key/response binding was randomized across subjects, so page 3 provided below only applied to half of our participants, where the other half had a corresponding, flipped association of key and responses.

  \begin{tcolorbox}[sharp corners, title=Page 1]
    \small
    \begin{itemize}
      \item This experiment is about sentence processing
      \item You will read sentences on the screen, with the words presented one after the other, at the center of the screen
      \item Some of these sentences will contain mistakes
      \item Your task is to find these mistakes
    \end{itemize}
  \end{tcolorbox}

  \begin{tcolorbox}[sharp corners, title=Page 2]
    \small
    Here are a few examples to show what we mean by correct and incorrect. Remember that the sentence will not be presented as a whole, but rather one word after another.

    Incorrect examples:
    \begin{itemize}
      \item The boy drink water while listening to music
      \item The farmer near the two pilot detests boxing
      \item The athletes that dislike the happy proud banker sings
    \end{itemize}

    Correct examples:

    \begin{itemize}
      \item The boy drinks water while listening to music
      \item The farmer near the two pilots detests boxing
      \item The athletes that dislike the happy proud banker sing
    \end{itemize}

    Some sentences might be a bit weird, like in example~3, but you should always be able to perform the task if you remain focused.
  \end{tcolorbox}

  \begin{tcolorbox}[sharp corners, title=Page 3]
    \small
    You have to look at the cross at the center of the screen, which is always present when there is no word to read. Make sure the luminosity of your screen is high enough for you to read. Then you will read sentences one word after the other and you have to do the following:

    \begin{itemize}
      \item As soon as you think a given sentence is INCORRECT, please press the -> right arrow key on your keyboard
      \item When the sentence ends, if you think it is CORRECT, please press the <- left arrow key on your keyboard
      \item You have to answer every time, even when you're not sure or you feel you don't know. Only after you answer, the following sentence will start. Answer the best you can!
    \end{itemize}

    After each answer you will receive feedback: the central cross will turn green if you answered correctly, and red otherwise. If you can, please turn your computer audio on: that way, you will receive feedback with sounds for each trial.

    This is the last instruction page. You can go back to the other pages, but when you move forward the experiment ask you to go fullscreen. Then the experiment will start with 5 training examples so that you understand the task.
  \end{tcolorbox}

  \section*{Material}

  All stimuli are provided below, first the grammatical ones (1-50), then the ungrammatical ones (51-100):

  \begin{spacing}{0.05}
    \setstretch{0.05}
    \begin{enumerate}[noitemsep]
      \footnotesize
      \item The lawyers who avoid the kind gentle judge lie.
      \item The athlete who hates the farmers the least sings.
      \item The judges that fear the proud charming man sing.
      \item The builder who dislikes the proud gentle farmer cheats.
      \item The plumbers that run happily although rather quickly lie.
      \item The painter who loves the young lazy farmers cheats.
      \item The waiter who avoids the judge the least cooks.
      \item The waiters that love the chefs the least pray.
      \item The tailor who avoids the farmer the least prays.
      \item The waiters that walk happily albeit pretty quickly pray.
      \item The judges that avoid the tailors the most swim.
      \item The women who love the happy clever chefs sing.
      \item The lawyer that runs carefully yet fairly quickly swims.
      \item The athlete that loves the vet the most lies.
      \item The waiters who walk carefully yet pretty quickly lie.
      \item The teacher who fears the lawyers the most cheats.
      \item The teachers who fear the plumber the most climb.
      \item The waiters who avoid the clumsy clever plumber cheat.
      \item The waiter who dislikes the proud nice woman swims.
      \item The chefs who avoid the waiters the least climb.
      \item The painters who fear the funny nice women pray.
      \item The vets that like the farmer the most smoke.
      \item The doctor that runs happily albeit rather carefully cheats.
      \item The teachers that dislike the tailors the least cheat.
      \item The lawyers that love the waiter the least climb.
      \item The plumber who fears the lawyer the most climbs.
      \item The painters that love the careless proud judges smoke.
      \item The farmers that run happily though fairly quickly sing.
      \item The waiters that walk carefully although rather quickly cheat.
      \item The man who laughs happily though pretty quickly lies.
      \item The plumber that rides happily although pretty quickly swims.
      \item The actor that dislikes the lawyer the most prays.
      \item The chef who dislikes the authors the least cheats.
      \item The vet that likes the proud helpful painters cheats.
      \item The farmer who fears the clever lazy tailors cheats.
      \item The painter who dislikes the nice careless teacher cheats.
      \item The painters that dislike the careless helpful tailors climb.
      \item The waiters who avoid the men the most sing.
      \item The actors that hate the painter the most cook.
      \item The teacher who likes the bakers the most sings.
      \item The actor who dislikes the chefs the most swims.
      \item The lawyers that fear the funny kind athlete pray.
      \item The bakers that fear the man the least pray.
      \item The actors who fear the nice proud men cook.
      \item The man who laughs carefully yet rather quickly smokes.
      \item The athlete who hates the proud funny woman prays.
      \item The tailor that loves the clever clumsy bakers cheats.
      \item The man who avoids the clumsy helpful chef sings.
      \item The vets who dislike the young nice man cheat.
      \item The man who fears the lazy nice authors lies.
      \item The baker who likes the clever happy plumber swims.
      \item The baker that hates the lazy gentle man cooks.
      \item The waiter who likes the actors the most smokes.
      \item The tailors who like the nice cool men swim.
      \item The women who fear the lazy clumsy plumber sing.
      \item The baker who dislikes the judge the least cooks.
      \item The woman that fears the baker the most cheats.
      \item The plumber who talks happily yet rather quickly swims.
      \item The lawyer that likes the farmers the most swims.
      \item The bakers that love the careless clever chef pray.
      \item The man who walks carefully although fairly quickly cheats.
      \item The man that runs carefully though rather quickly lies.
      \item The bakers who love the actor the least cheat.
      \item The tailor that dislikes the cool lazy baker prays.
      \item The plumbers who love the kind charming doctor cheat.
      \item The pilots that hate the waiter the most cook.
      \item The baker who dislikes the kind helpful plumbers lies.
      \item The farmer who fears the doctors the least swims.
      \item The doctor who hates the actor the least cheats.
      \item The waiter who hates the cool clumsy man swims.
      \item The farmer who likes the builders the least prays.
      \item The waiter who dislikes the painter the least prays.
      \item The woman who avoids the gentle lazy waiters prays.
      \item The author that hates the waiters the least smokes.
      \item The doctor that hates the careless young teacher lies.
      \item The chef that dislikes the proud clumsy tailors sings.
      \item The athletes who love the judges the least sing.
      \item The bakers that love the lovely helpful women cook.
      \item The tailors that drive happily although fairly quickly climb.
      \item The tailors who run carefully yet fairly happily cheat.
      \item The chefs who love the proud cool bakers cook.
      \item The bakers who fear the woman the most climb.
      \item The teacher who dislikes the helpful charming builders cheats.
      \item The chefs who run carefully albeit rather quickly cook.
      \item The painter who loves the helpful friendly judges prays.
      \item The waiters that run carefully though pretty quickly cook.
      \item The teachers that avoid the waiters the most pray.
      \item The waiters who drive happily yet fairly quickly swim.
      \item The painter that avoids the waiter the most prays.
      \item The tailors that love the charming young authors smoke.
      \item The lawyers that dislike the proud young farmer cheat.
      \item The vets that love the woman the least cook.
      \item The doctors that like the bakers the least pray.
      \item The builder that drives happily though rather quickly cheats.
      \item The plumbers who dislike the careless clever bakers sing.
      \item The athletes who hate the bakers the most swim.
      \item The doctors that like the chef the most cook.
      \item The bakers who fear the friendly nice man swim.
      \item The men who dislike the pilots the least pray.
      \item The plumber that laughs carefully yet pretty quickly prays.
    \end{enumerate}
  \end{spacing}

  \newpage
  \section{Additional Results}
  \label{appendix:appendixB}

  \begin{table}[H]
    \caption{ANOVAs for two dependent variables in humans (response time and error rates) and for error rates in T5, testing whether there was a significant effect of violation in three possible conditions: Baseline, (no attractor), and Distal and Proximal attractors.}
    \centering
    \begin{tabular}[t]{llll}
      \toprule
      condition                                           & Statistic                                         & $p$-value                              & $\eta_G^2$                            \\
      \midrule
      \addlinespace[0.3em]
      \multicolumn{4}{l}{\textbf{Response Time (humans)}}                                                                                                                                      \\
      \hspace{1em}Baseline                                & $F_{1,33}=1.81$                                   & $.187$                                 & $.023$                                \\
      \cellcolor[HTML]{fff3e0}{\em{\hspace{1em}Distal}}   & \cellcolor[HTML]{fff3e0}{\em{$F_{1,33}=27.86$}}   & \cellcolor[HTML]{fff3e0}{\em{$<.001$}} & \cellcolor[HTML]{fff3e0}{\em{$.274$}} \\
      \cellcolor[HTML]{fff3e0}{\em{\hspace{1em}Proximal}} & \cellcolor[HTML]{fff3e0}{\em{$F_{1,33}=13.22$}}   & \cellcolor[HTML]{fff3e0}{\em{$<.001$}} & \cellcolor[HTML]{fff3e0}{\em{$.139$}} \\
      \addlinespace[0.3em]
      \multicolumn{4}{l}{\textbf{Error Rate (humans)}}                                                                                                                                         \\
      \cellcolor[HTML]{fff3e0}{\em{\hspace{1em}Baseline}} & \cellcolor[HTML]{fff3e0}{\em{$F_{1,33}=10.91$}}   & \cellcolor[HTML]{fff3e0}{\em{$.002$}}  & \cellcolor[HTML]{fff3e0}{\em{$.040$}} \\
      \cellcolor[HTML]{fff3e0}{\em{\hspace{1em}Distal}}   & \cellcolor[HTML]{fff3e0}{\em{$F_{1,33}=22.76$}}   & \cellcolor[HTML]{fff3e0}{\em{$<.001$}} & \cellcolor[HTML]{fff3e0}{\em{$.032$}} \\
      \cellcolor[HTML]{fff3e0}{\em{\hspace{1em}Proximal}} & \cellcolor[HTML]{fff3e0}{\em{$F_{1,33}=28.60$}}   & \cellcolor[HTML]{fff3e0}{\em{$<.001$}} & \cellcolor[HTML]{fff3e0}{\em{$.047$}} \\
      \addlinespace[0.3em]
      \multicolumn{4}{l}{\textbf{Error Rate (T5)}}                                                                                                                                             \\
      \cellcolor[HTML]{fff3e0}{\em{\hspace{1em}Baseline}} & \cellcolor[HTML]{fff3e0}{\em{$F_{1,110}=6.29$}}   & \cellcolor[HTML]{fff3e0}{\em{$.014$}}  & \cellcolor[HTML]{fff3e0}{\em{$.054$}} \\
      \cellcolor[HTML]{fff3e0}{\em{\hspace{1em}Distal}}   & \cellcolor[HTML]{fff3e0}{\em{$F_{1,233}=71.59$}}  & \cellcolor[HTML]{fff3e0}{\em{$<.001$}} & \cellcolor[HTML]{fff3e0}{\em{$.235$}} \\
      \cellcolor[HTML]{fff3e0}{\em{\hspace{1em}Proximal}} & \cellcolor[HTML]{fff3e0}{\em{$F_{1,216}=147.35$}} & \cellcolor[HTML]{fff3e0}{\em{$<.001$}} & \cellcolor[HTML]{fff3e0}{\em{$.406$}} \\
      \bottomrule
    \end{tabular}

  \end{table}

  \begin{table}[H]
    \caption{Corresponding table to Table~\ref{tab:table1} with the filter operation when the attractor is singular.}
    \centering
    \begin{tabular}[t]{llll}
      \toprule
      Effect                                               & $F_{1,33}$                             & $p$-value                              & $\eta_G^2$                            \\
      \midrule
      \addlinespace[0.3em]
      \multicolumn{4}{l}{\textbf{Response Time; Distal attractor}}                                                                                                                   \\
      \hspace{1em}congruency                               & $1.12$                                 & $.297$                                 & $.001$                                \\
      \cellcolor[HTML]{fff3e0}{\em{\hspace{1em}violation}} & \cellcolor[HTML]{fff3e0}{\em{$14.02$}} & \cellcolor[HTML]{fff3e0}{\em{$<.001$}} & \cellcolor[HTML]{fff3e0}{\em{$.021$}} \\
      \hspace{1em}interaction                              & $0.15$                                 & $.704$                                 & $<.001$                               \\
      \addlinespace[0.3em]
      \multicolumn{4}{l}{\textbf{Response Time; Proximal attractor}}                                                                                                                 \\
      \hspace{1em}congruency                               & $0.00$                                 & $.961$                                 & $<.001$                               \\
      \cellcolor[HTML]{fff3e0}{\em{\hspace{1em}violation}} & \cellcolor[HTML]{fff3e0}{\em{$13.66$}} & \cellcolor[HTML]{fff3e0}{\em{$<.001$}} & \cellcolor[HTML]{fff3e0}{\em{$.045$}} \\
      \hspace{1em}interaction                              & $0.32$                                 & $.578$                                 & $<.001$                               \\
      \addlinespace[0.3em]
      \multicolumn{4}{l}{\textbf{Error Rate; Distal attractor}}                                                                                                                      \\
      \hspace{1em}congruency                               & $0.03$                                 & $.867$                                 & $<.001$                               \\
      \cellcolor[HTML]{fff3e0}{\em{\hspace{1em}violation}} & \cellcolor[HTML]{fff3e0}{\em{$14.52$}} & \cellcolor[HTML]{fff3e0}{\em{$<.001$}} & \cellcolor[HTML]{fff3e0}{\em{$.091$}} \\
      \hspace{1em}interaction                              & $0.88$                                 & $.356$                                 & $.003$                                \\
      \addlinespace[0.3em]
      \multicolumn{4}{l}{\textbf{Error Rate; Proximal attractor}}                                                                                                                    \\
      \hspace{1em}congruency                               & $0.00$                                 & $>.999$                                & $<.001$                               \\
      \cellcolor[HTML]{fff3e0}{\em{\hspace{1em}violation}} & \cellcolor[HTML]{fff3e0}{\em{$8.17$}}  & \cellcolor[HTML]{fff3e0}{\em{$.007$}}  & \cellcolor[HTML]{fff3e0}{\em{$.053$}} \\
      \hspace{1em}interaction                              & $0.38$                                 & $.539$                                 & $.002$                                \\
      \bottomrule
    \end{tabular}
    \label{tab:table2singular}
  \end{table}

  \begin{figure*}
    \centering
    \includegraphics[width=.85\textwidth]{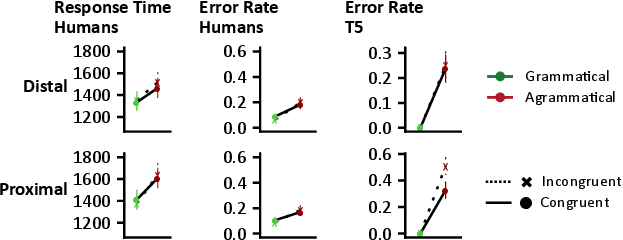}
    \caption{Corresponding figure to \ref{fig:detailedResults} filtering for singular attractors: observe the overall collapse of the effect of congruency.}
    \label{fig:detailedResultsAppendix}
  \end{figure*}

  \begin{table}

    \caption{Table~\ref{tab:table2} with full dataset of N=54 participants included}
    \centering
    \begin{tabular}[t]{llll}
      \toprule
      Effect                                                & $F_{1,53}$                             & $p$-value                              & $\eta_G^2$                             \\
      \midrule
      \addlinespace[0.3em]
      \multicolumn{4}{l}{\textbf{Response Time; Distal attractor}}                                                                                                                     \\
      \cellcolor[HTML]{fff3e0}{\em{\hspace{1em}congruency}} & \cellcolor[HTML]{fff3e0}{\em{$0.14$}}  & \cellcolor[HTML]{fff3e0}{\em{$.714$}}  & \cellcolor[HTML]{fff3e0}{\em{$<.001$}} \\
      \cellcolor[HTML]{fff3e0}{\em{\hspace{1em}violation}}  & \cellcolor[HTML]{fff3e0}{\em{$11.77$}} & \cellcolor[HTML]{fff3e0}{\em{$.001$}}  & \cellcolor[HTML]{fff3e0}{\em{$.024$}}  \\
      \hspace{1em}interaction                               & $0.08$                                 & $.773$                                 & $<.001$                                \\
      \addlinespace[0.3em]
      \multicolumn{4}{l}{\textbf{Response Time; Proximal attractor}}                                                                                                                   \\
      \cellcolor[HTML]{fff3e0}{\em{\hspace{1em}congruency}} & \cellcolor[HTML]{fff3e0}{\em{$0.16$}}  & \cellcolor[HTML]{fff3e0}{\em{$.691$}}  & \cellcolor[HTML]{fff3e0}{\em{$<.001$}} \\
      \cellcolor[HTML]{fff3e0}{\em{\hspace{1em}violation}}  & \cellcolor[HTML]{fff3e0}{\em{$17.98$}} & \cellcolor[HTML]{fff3e0}{\em{$<.001$}} & \cellcolor[HTML]{fff3e0}{\em{$.038$}}  \\
      \hspace{1em}interaction                               & $0.97$                                 & $.329$                                 & $.002$                                 \\
      \addlinespace[0.3em]
      \multicolumn{4}{l}{\textbf{Error Rate; Distal attractor}}                                                                                                                        \\
      \hspace{1em}congruency                                & $0.01$                                 & $.918$                                 & $<.001$                                \\
      \cellcolor[HTML]{fff3e0}{\em{\hspace{1em}violation}}  & \cellcolor[HTML]{fff3e0}{\em{$20.70$}} & \cellcolor[HTML]{fff3e0}{\em{$<.001$}} & \cellcolor[HTML]{fff3e0}{\em{$.062$}}  \\
      \hspace{1em}interaction                               & $0.09$                                 & $.768$                                 & $<.001$                                \\
      \addlinespace[0.3em]
      \multicolumn{4}{l}{\textbf{Error Rate; Proximal attractor}}                                                                                                                      \\
      \cellcolor[HTML]{fff3e0}{\em{\hspace{1em}congruency}} & \cellcolor[HTML]{fff3e0}{\em{$0.75$}}  & \cellcolor[HTML]{fff3e0}{\em{$.391$}}  & \cellcolor[HTML]{fff3e0}{\em{$.002$}}  \\
      \cellcolor[HTML]{fff3e0}{\em{\hspace{1em}violation}}  & \cellcolor[HTML]{fff3e0}{\em{$18.21$}} & \cellcolor[HTML]{fff3e0}{\em{$<.001$}} & \cellcolor[HTML]{fff3e0}{\em{$.064$}}  \\
      \hspace{1em}interaction                               & $0.27$                                 & $.607$                                 & $<.001$                                \\
      \bottomrule
    \end{tabular}
    \label{tab:table2fulldataset}
  \end{table}

  \begin{figure}
    \centering
    \includegraphics{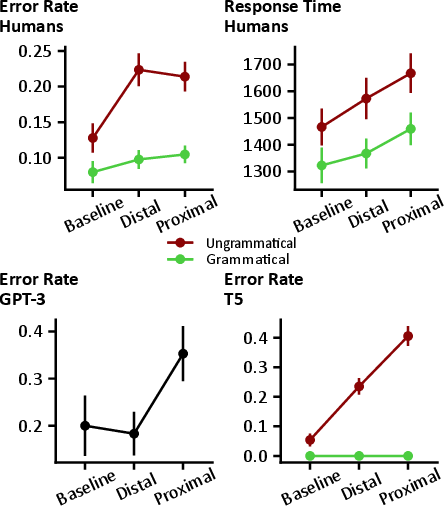}
    \caption{Performances of humans and NLMs, using all participants: colors
      indicate grammaticality, error bars indicate (all figures)
      SEM over participants (humans) or sentences (NLMs). Results are comparable
      to the ones after rejection of low-performing participants}
    \label{fig:fig2allparticipants}
  \end{figure}

\end{appendices}

\end{document}